\documentclass[10pt,journal,compsoc]{IEEEtran}
\ifCLASSOPTIONcompsoc
  \usepackage[nocompress]{cite}
\else
  \usepackage{cite}
\fi
\ifCLASSINFOpdf
\else
\fi

\usepackage{threeparttable}

\usepackage{multirow}
\usepackage{graphicx}
\usepackage{amsmath}
\usepackage{subfigure}
\usepackage{multirow}
\usepackage[square, comma, sort&compress, numbers]{natbib}

\begin{document}
%
\title{AST-GCN: Attribute-Augmented Spatiotemporal Graph Convolutional Network for Traffic Forecasting}
%
%
%
%

\author{Jiawei Zhu, Chao Tao, Hanhan Deng, Ling Zhao, Pu Wang,~\IEEEmembership{Member,~IEEE,} Tao Lin, Haifeng Li*,~\IEEEmembership{Member,~IEEE,}
\IEEEcompsocitemizethanks{\IEEEcompsocthanksitem H. Li, J. Zhu, C. Tao and L. Zhao are with School of Geosciences and Info-Physics, Central South University, Changsha 410083, Hunan, China.
(e-mail:
lihaifeng@csu.edu.cn)
\IEEEcompsocthanksitem H. Deng  is with Hisense TransTech Co., Ltd., No.17 Donghai West Road, Qingdao, China
\IEEEcompsocthanksitem P. Wang is with with School of Traffic Transportation Engineering, Central South University, Changsha 410083, China.
\IEEEcompsocthanksitem T. Lin is with College of Biosystems Engineering and Food Science, Zhejiang University, Hangzhou, China.}
\thanks{Corresponding author: Haifeng Li.\\
\textcopyright 2020 IEEE}}

%
%

\markboth{Submitted to IEEE Transactions on Intelligent Transportation Systems}%
{}
%



\IEEEtitleabstractindextext{%
\begin{abstract}
  Traffic forecasting is a fundamental and challenging task in the field of intelligent transportation. Accurate forecasting not only depends on the historical traffic flow information but also needs to consider the influence of a variety of external factors, such as weather conditions and surrounding POI distribution. Recently, spatiotemporal models integrating graph convolutional networks and recurrent neural networks have become traffic forecasting research hotspots and have made significant progress. However, few works integrate external factors. Therefore, based on the assumption that introducing external factors can enhance the spatiotemporal accuracy in predicting traffic and improving interpretability, we propose an attribute-augmented spatiotemporal graph convolutional network (AST-GCN). We model the external factors as dynamic attributes and static attributes and design an attribute-augmented unit to encode and integrate those factors into the spatiotemporal graph convolution model. Experiments on real datasets show the effectiveness of considering external information on traffic forecasting tasks when compared to traditional traffic prediction methods. Moreover, under different attribute-augmented schemes and prediction horizon settings, the forecasting accuracy of the AST-GCN is higher than that of the baselines. 
\end{abstract}

}

\maketitle

\IEEEdisplaynontitleabstractindextext

%
\IEEEpeerreviewmaketitle

\IEEEraisesectionheading{\section{Introduction}\label{sec:introduction}}

%
%
%
%
\IEEEPARstart{A}{s} one of the essential components in intelligent transportation systems (ITSs), traffic forecasting can provide a scientific basis for the management and planning of urban transportation systems\cite{zhang2011data,cui2019traffic, zheng2019deepstd}. According to predicted traffic states, transportation departments can deploy and guide traffic flows in advance, thereby improving the operating efficiency of road networks and alleviating traffic jams\cite{yu2017spatio,park2011real}.

The starting point of traditional traffic forecasting methods is generally to learn historical traffic characteristics to predict traffic at future moments\cite{yao2019revisiting}. However, it is difficult to achieve accurate traffic forecasting because future traffic states not only depend on historical states but can also be affected by a variety of static and dynamic external factors. Among them, the static factors will not change continuously over time, but the traffic state at a certain time will be affected. For example, on a road section with a large number of restaurants, the traffic state during the dining period will be significantly different from other periods; the latter will change over time and lead to changes in traffic conditions. Taking weather conditions as an example, as the weather shifts from sunny to heavy rain, the traffic speed will generally decrease. These factors create randomness in traffic states and make it challenging for accurate traffic forecasting.

Aiming at the problem that traditional models cannot comprehensively consider factors affecting traffic conditions, we propose an attribute-augmented spatiotemporal graph convolutional model (AST-GCN). We consider the external factors as attributes of road sections in the road network and model the attributes and traffic features of road sections simultaneously to obtain the augmented feature vectors. By this means, the model’s perception of the external information is enhanced, thereby improving forecasting accuracy.

The main contributions of this article are as follows:

(1) This paper proposes a novel traffic prediction model AST-GCN that can capture external information in combination with the spatiotemporal graph convolution model.

(2) The proposed AST-GCN model can integrate both dynamic and static external information related to the road, such as weather and surrounding POIs.;

(3) We evaluate the model using real data, and the experimental results show that the prediction results of the AST-GCN model outperform the baselines, indicating the effectiveness of modeling external information.

The remaining sections of this article are organized as follows. The second section reviews related works and development trends of traffic forecasting. Section 3 introduces the details of our method. In section 4, experiments are conducted on real-world data to evaluate the performance of the proposed method compared with baselines, and a perturbation analysis is carried out to test the robustness of our model. The summary and conclusion of future works are given in section 5.

\section{Related works}

Traffic forecasting is an essential part of ITS and plays an important role in urban traffic control and development. The traffic forecasting methods have undergone different stages of evolution. The traditional analysis is mainly based on mathematical statistics to predict traffic states at the beginning. Among them, the principle of the historical average model (HAM) uses historical average data as the prediction result, which is simple to calculate but has low prediction accuracy\cite{smith1997traffic}. Time series models such as ARMA\cite{ahmed1979analysis} and its variants\cite{williams1998urban,lee1999application} utilize the relationship between current data and historical data for forecasting and perform modeling and analysis considering the periodicity and trend of data. However, they are based on the time series stability assumption and are thus unable to capture traffic flow mutations. Later, traffic forecasting models based on machine learning emerged. The k-nearest neighbor algorithm (KNN) was first applied to the prediction of traffic flow \cite{davis1991nonparametric}, followed by studies using the Bayesian inference method\cite{tebaldi1998bayesian}, and support vector machine (SVM)\cite{hu2016short}.

These algorithms can model more complex characteristics of traffic flows but have limited ability to capture nonlinear patterns. In recent years, deep learning has attracted the attention of researchers due to its advantages in capturing nonlinear and complex patterns. Many deep learning methods have been applied to traffic forecasting, such as deep confidence networks (DBNs)\cite{huang2014deep,jia2016traffic} and stacked autoencoding neural networks (SAEs)\cite{lv2014traffic}. One of the disadvantages of these methods is that they independently process the traffic flow information at each time and do not directly model the dependencies in traffic flows in the time series. Therefore, using recurrent neural networks (RNNs) based on sequence prediction to predict traffic flow was studied. However, recurrent neural networks suffer from short-term memory due to the vanishing gradient during backpropagation. Therefore, researchers turned to long short-term memory networks (LSTMs)\cite{tian2015predicting,ma2015long} and gated recurrent units (GRUs)\cite{fu2016using,zhang2018combining}, which were created as solutions to short-term memory to extract the temporal dependencies of traffic flow data.

Although these models can capture temporal dependencies in traffic flows, researchers have gradually recognized the importance of spatial dependencies and have made improvements by introducing convolutional neural networks (CNNs) to extract spatial information and combining them with LSTMs\cite{liu2017short,wu2016short}, which has improved prediction accuracy. Since CNNs were designed for Euclidean space, such as images and grids, they have limitations in transportation networks with non-Euclidean topology and thus cannot essentially characterize the spatial dependence of traffic flows. Emerging graph convolutional neural networks (GCNs) are dedicated to processing network structures\cite{kipf2016semi,zhou2018graph}, which can better model the spatial dependence of road segments on traffic networks\cite{zhao2019t,li2019hybrid}.

However, the traffic prediction task not only relies on historical traffic information and spatial relationships but is also affected by a variety of external factors, such as weather conditions and the distribution of surrounding POIs. How to integrate the information about external influence factors in the model is the main problem of the current traffic task. In previous studies, there have been considerations for multisource data. For example, Liao et al.\cite{liao2018deep} integrated an encoder based on LSTM \cite{van2002freeway} to encode external information and model multimodal data as a sequence input. The model proposed by Zhang et al.\cite{zhang2018combining} is mainly based on the GRU model\cite{fu2016using} and realizes the traffic forecasting task with external weather information based on the feature fusion of the input features and weather information.

In summary, the existing methods do not fully consider the impact of internal and external factors simultaneously. How to combine multisource data to realize the task of traffic prediction is an urgent problem to be solved. Therefore, this paper proposes an attribute-augmented spatiotemporal graph convolutional model (AST-GCN) for traffic forecasting, which regards external factors as attribute information of the road segments in the road network. The attribute information and traffic characteristics are then integrated to enhance the model's perception of external information, thereby improving traffic forecasting accuracy.

\section{Method}
\subsection{Problem Definition}

The goal of traffic forecasting is to predict future traffic states based on historical states and auxiliary information. While the traffic states of road sections are mainly described by the average traffic volume, speed, and occupancy rate, we take the average traffic speed as an example to illustrate our work. Therefore, the traffic forecasting task of this paper is mainly based on the traffic speeds in the past period of time and external factors that affect the traffic to predict the traffic speed in the future.

Definition 1: Road network $G$. We use a road network $G=(V,E)$ to represent the connection relationship between road sections. $V={v_1,v_2,\dots,v_n}$ represents the collection of road sections, n is the number of road sections, $E={e_1,e_2,\dots,e_m}$ is the collection of edges that indicate the connectivity between two road segments, and m represents the number of edges. Without loss of generality, the adjacency matrix $A$ is used to illustrate the connectivity of the road network. $A$ is a matrix composed of 0 and 1 when $G$ is an unweighted network, where 1 suggests that corresponding road segments are connected, and 0 otherwise.

Definition 2: Traffic feature matrix $X$. In this paper, the traffic speed is regarded as an inherent attribute of each node on the urban road network, represented by a matrix $X$, while $x^t_{i}$ represents the traffic speed on the i-th road section at time t.

Definition 3: Attribute matrix $K$. In this paper, the external factors that affect traffic conditions are regarded as the auxiliary attributes of the road segments on the urban road network. These factors can form an attribute matrix $K=\left\{K_{1}, K_{2}, \ldots, K_{l}\right\}$, where $l$ is the category number of auxiliary information. The set of auxiliary information of type $j$ is represented as $K_{j}=\left\{j^{1}, j^{2}, \ldots, j^{t}\right\}$, and $j_i^t$ is the j-th auxiliary information of the ith road section at time t.

In summary, the traffic prediction problem can be understood as learning the function $f$ on the basis of the basic topology $G$, feature matrix $X$ and attribute matrix $K$ of the road network to obtain traffic information in the future period T, as shown in Eq.\ref{equa:1}:

\begin{equation}
 \left[x_{t+1}, x_{t+2}, \ldots, x_{t+T}\right]=f(G, {X} \mid K) \label{equa:1}
\end{equation}

\subsection{External factors}
\begin{figure}[h]
	\centering
	   \includegraphics[width=1.0\linewidth]{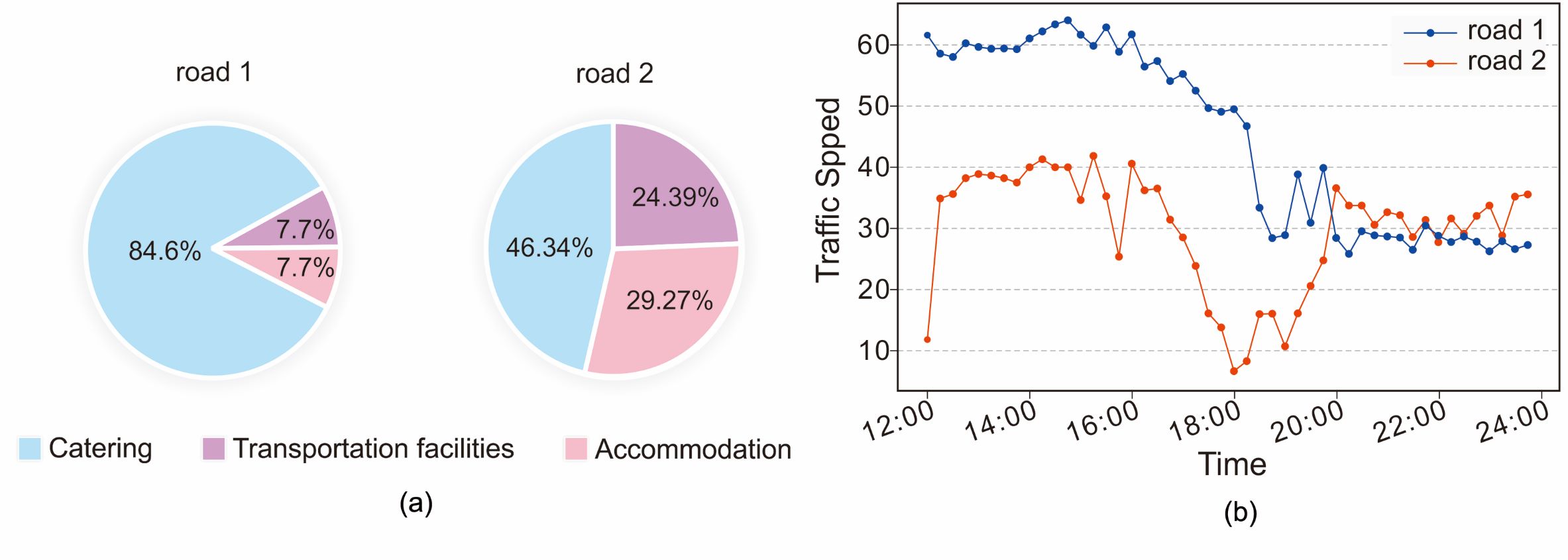}
	   \caption{Illustration of the influence of POIs on traffic states.}
	   \label{fig:1}
\end{figure}

Without loss of generality, this article analyzes the impact of external factors on traffic states from static and dynamic perspectives:

(1) Static factors, which mainly refer to static geographic information that does not change with time but still exerts influence on traffic states. For example, the distribution of POIs around a road section can determine the visiting pattern of people and the attractiveness of the road section, which is reflected by its traffic states in return. Fig.\ref{fig:1}-(a) shows the distribution of three different kinds of POIs (catering, transportation, and accommodation) around two road sections. Fig. \ref{fig:1}-b shows the variation in traffic speed on two road sections on the same day. Road 1, with more restaurants and fewer traffic facilities, has a higher traffic speed than road 2 between 12:00-20:00, which validates that the difference in the distribution of POIs will lead to a difference in traffic states.

\begin{figure}[h]
	\centering
	   \includegraphics[width=0.5\linewidth]{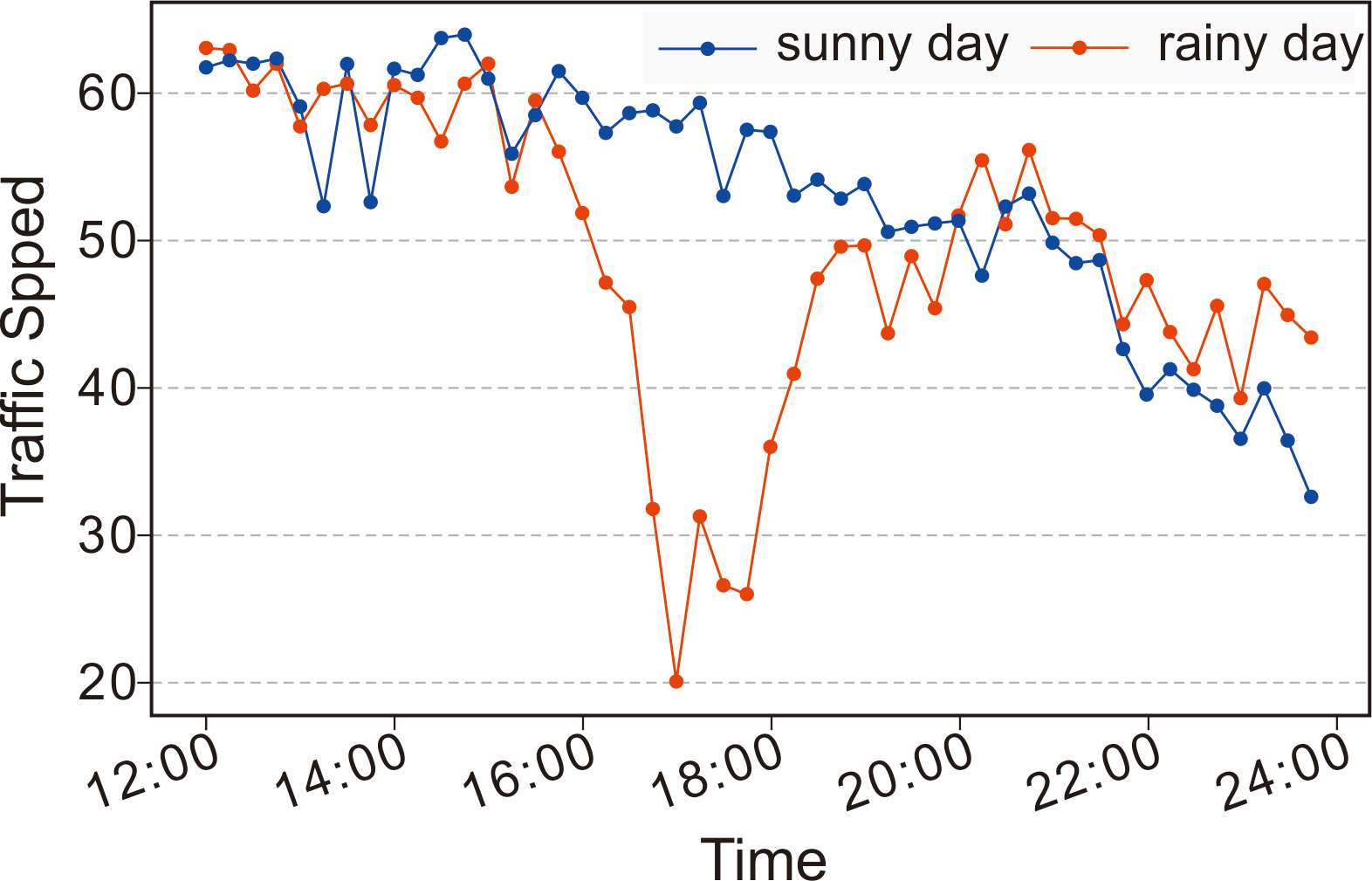}
	   \caption{Illustration of the influence of weather conditions on traffic states.}
	   \label{fig:2}
\end{figure}

(2) Dynamic factors, such as the weather condition, which is a time-varying determinant of road conditions and visions while driving, can directly affect the traffic states. As an illustration of the influence of weather conditions, Fig.\ref{fig:2} shows the change in traffic speed over time of a certain road section on separate days with different weather conditions. Specifically, one day has heavy rain from 16 o'clock to 18 o'clock and then turns into light rain; the other day is sunny all day. It can be seen that the traffic speed decreased greatly during the rainy hours compared to the counterpart on a sunny day, which indicates the great impact of weather on traffic states.

\subsection{Attribute augmentation unit}
To comprehensively consider factors that affect the traffic states, this study models external factors as the dynamic ($D$) and static ($S$) attributes of the road segments in the road network. Then, the traffic feature matrix $X$ and the attribute matrix $K=\{S,D\}$ are fed into the attribute augmentation unit (A-Cell) to derive augmented matrices.

(1) Incorporating static attributes $S$.

$S\in R^{n\times p}$ is a collection of $p$ different static attributes $\{\vec{s_1},\vec{s_2},\ldots,\vec{s_p}\}$. Since the attribute values do not vary with time, the whole matrix $S$ is constantly used, while only the corresponding column of feature matrix $X$ is extracted in the process of generating the augmented matrix at each timestamp. Specifically, the matrix augmented by static attributes at time t is formed as:

\begin{equation}
E^t_s=[X^t, S], \   E^t_s\in R^{n\times (p+l)}
  \label{equa:2}
 \end{equation}

(2) Incorporating dynamic attributes $D$.

Different from $S$, $D\in R^{n\times (w*t)}$ is a collection of $w$ different dynamic attributes $\{D_1,D_2,\ldots,D_w\}$. Notably, considering that the traffic states are subject to the cumulative effects of dynamic factors over a period, instead of selecting attribute values corresponding to time t, we expand the selecting window size to $m+1$ when forming $E^t$, that is, selecting $D_w^{t-m,t}=[D_w^{t-m},D_w^{t-m-1},\ldots, D_w^{t}]$ for each dynamic attribute submatrix $D_w$. Finally, through the attribute augmentation unit (A-Cell), the augmented matrix containing both static and dynamic external attributes as well as traffic characteristic information at time $t$ is formed as:

\begin{equation}
  \begin{aligned}
    E^t =[X^t, S, D_1^{t-m,t},D_2^{t-m,t},\ldots,D_w^{t-m,t}]
    \end{aligned}
  \label{equa:3}
 \end{equation}

where $E^t\in R^{n\times (p+l+w*(m+1))}$.





\subsection{Spatiotemporal graph convolutional network}
Road networks are naturally graph structures. Since traffic flows on connected road sections, modeling spatial dependencies on the network is essential in the traffic forecasting task. Graph convolutional networks (GCNs)\cite{kipf2016semi} have recently been favored by many researchers because of their ability to model complex relationships and interdependency in non-Euclidean domains and have made great progress in recent years. The main purpose of the graph convolutional neural network is to obtain the representation of each node in the graph while considering the influence of neighboring nodes. To capture the dependency created by topological structures, GCN takes the adjacency matrix and the feature matrix as inputs, and the modeling process can be expressed as follows.

\begin{equation}
  \begin{aligned}
    y_{l+1}=\sigma\left(\tilde{D}^{-\frac{1}{2}} \tilde{A} \tilde{D}^{-\frac{1}{2}} y_{l} W_{l}\right)
    \end{aligned}
  \label{equa:4}
 \end{equation}
where $\sigma$ is the activation function, $\tilde{A}=A+I$ represents the adjacency matrix with self-loops, $\tilde{D}$ is the corresponding degree matrix, $W_{l}$ is the weight matrix of the $l$-th convolutional layer, $y_l$ is the output representation, and $y_0=X$.

\begin{figure}[ht]
  \centering
     \includegraphics[width=1\linewidth]{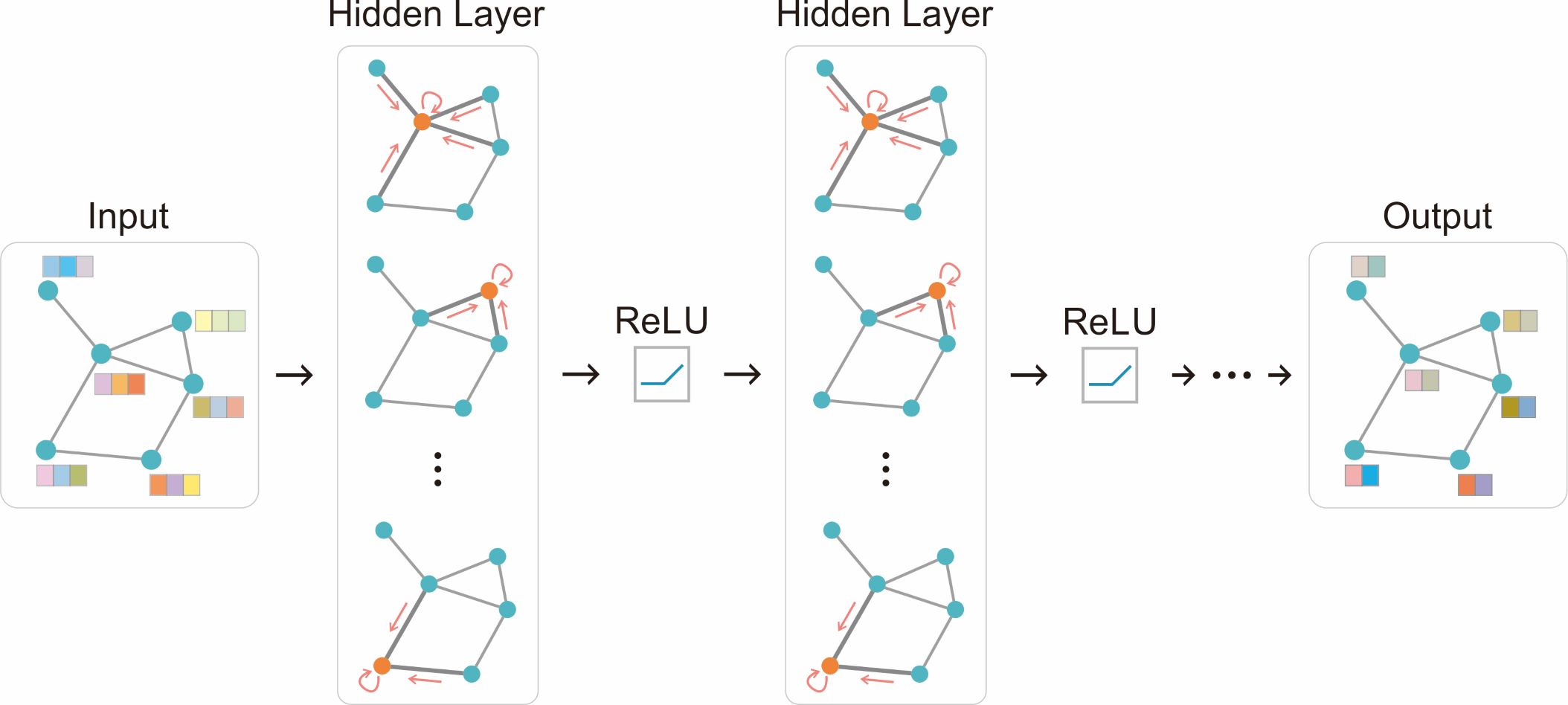}
     \caption{The architecture of the GCN model.}
     \label{fig:3}
\end{figure}

\begin{figure}[ht]
	\centering
     \includegraphics[width=1\linewidth]{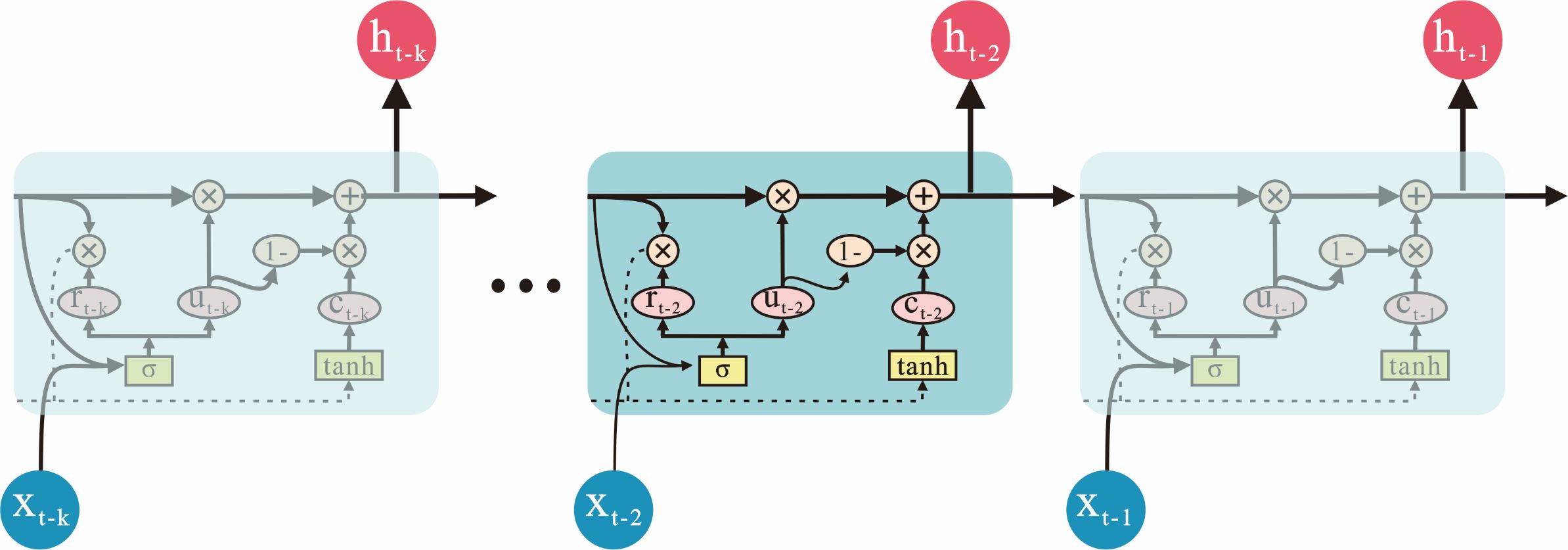}
     \caption{The architecture of the GRU model.}
     \label{fig:4}
\end{figure}

Moreover, traffic states can constantly change over time in the real world. To capture the temporal dependencies in traffic time series, recurrent neural networks\cite{connor1994recurrent} (RNNs) are introduced. However, traditional RNNs have defects such as the inability to retain long-term memory and the existence of exploding gradient. LSTM\cite{hochreiter1997long}, a variant of RNN, provides a solution to such problems. It designs the input gate and the forget gate unit to retain and forget information of the previous state and current input and derive the current state with the output gate unit. Based on LSTM, GRU\cite{cho2014properties} replaces the forget gate and input gate with one update gate, which both reduces the training time and improves the training efficiency while having equivalent calculation accuracy. Specifically, the GRU model can be regarded as compositions of reset gates and update gates, as shown in Fig. \ref{fig:4}, where $x_{t-1}$ is the input feature of one node at time $t-1$ and $h_{t-k},\ldots,h_{t-1},h_t$ represent the hidden states at time ${t-k},\ldots,{t-1},t$.  $\sigma$ and $tanh$ refer to the sigmoid and tanh activation, indicating gate signals. Consider gates at $t$ for example, $r_t$ is the reset gate used to combine previous state $h_{t-1}$ with current information $x_t$ to derive the candidate hidden state $c_t$. $u_t$ refers to the update gate that can be used to determine how much of $h_{t-1}$ to discard and what new information of $c_t$ to incorporate to derive the final hidden state $h_t$.

By using the combination of the GCN and GRU model, the purpose of capturing the spatiotemporal dependencies in real-world traffic data can be achieved. The GCN model is used to generate representations of road sections that capture the spatial dependencies on the road network at each timestamp. Then, these time-varying representations are fed into the GRU model to capture the temporal dependencies.


\subsubsection{Attribute-enhanced spatiotemporal graph convolution model (AST-GCN)}

Based on the spatiotemporal graph convolutional network, we propose a traffic forecasting model AST-GCN that integrates the information of external factors with the attribute augmentation units.

\begin{figure}[ht]
	\centering
     \includegraphics[width=0.8\linewidth]{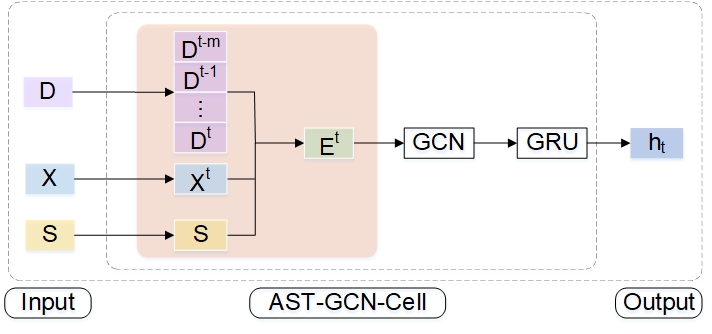}
     \caption{The architecture of an AST-GCN cell.}
     \label{fig:5}
\end{figure}

As shown in Fig. \ref{fig:5}, the attribute enhancement unit expands the dimension of the original feature matrix by incorporating static and dynamic external attributes. At time $t$, $X^t$ is extracted from the original traffic feature matrix $X$, $\{D^{t-m},\ldots,D^t\}$ is the collection of dynamic information from time $t-m$ to $t$, $S$ is the static attribute, which is constantly the same for different timestamps, and $E^t$ is the enhanced matrix after the fusion of the external attributes and original traffic features.

The enhancement matrix $E$ is then used as the input of the spatiotemporal model $f$ to obtain the final prediction result $y$.

\begin{equation}
  \begin{aligned}
    \hat{y}=f(A,X,E)
    \end{aligned}
  \label{equa:5}
 \end{equation}

 Specifically, the enhanced matrices are fed into a series of GCNs to generate time-varying features of road sections that encode the spatial characteristics of the traffic states. Then, the feature series are used as the input of GRUs to model the temporal dependencies and derive hidden traffic states.

\begin{equation}
  \begin{aligned}
    u_{t}=\sigma\left(W_{u} \cdot\left[g c\left(E^t, A\right), h_{t-1}\right]+b_{u}\right)
    \end{aligned}
  \label{equa:6}
 \end{equation}

 \begin{equation}
  \begin{aligned}
    r_{t}=\sigma\left(W_{r} \cdot\left[g c\left(E^t, A\right), h_{t-1}\right]+b_{r}\right)
    \end{aligned}
  \label{equa:7}
 \end{equation}

 \begin{equation}
  \begin{aligned}
    c_{t}=\tanh \left(W_{c} \cdot\left[g c\left(E^t, A\right),\left(r_{t}, h_{t-1}\right)\right]+b_{c}\right)
    \end{aligned}
  \label{equa:8}
 \end{equation}

 \begin{equation}
  \begin{aligned}
    h_{t}=u_{t} * h_{t-1}+\left(1-u_{t}\right) * c_{t}
    \end{aligned}
  \label{equa:9}
 \end{equation}
where $gc(\cdot)$ represents the graph convolution operation.




\subsubsection{Loss Function}

In the process of model training, the goal of traffic forecasting is to make the prediction result approximate the real traffic states as much as possible. Therefore, the objective of the loss function is to minimize the prediction error.
\begin{equation}
  \begin{aligned}
    {Loss}=\left\|y_{t}-\hat{y}_{t}\right\|+\lambda L_{r e g}
      \end{aligned}
  \label{equa:10}
 \end{equation}
where $y_t$ and $\hat{y_t}$ are the ground truth and prediction, $L_{r e g}$ represents the regular term to avoid overfitting, and $\lambda$ is a hyperparameter.

\subsection{Framework}

\begin{figure}[ht]
	\centering
     \includegraphics[width=1\linewidth]{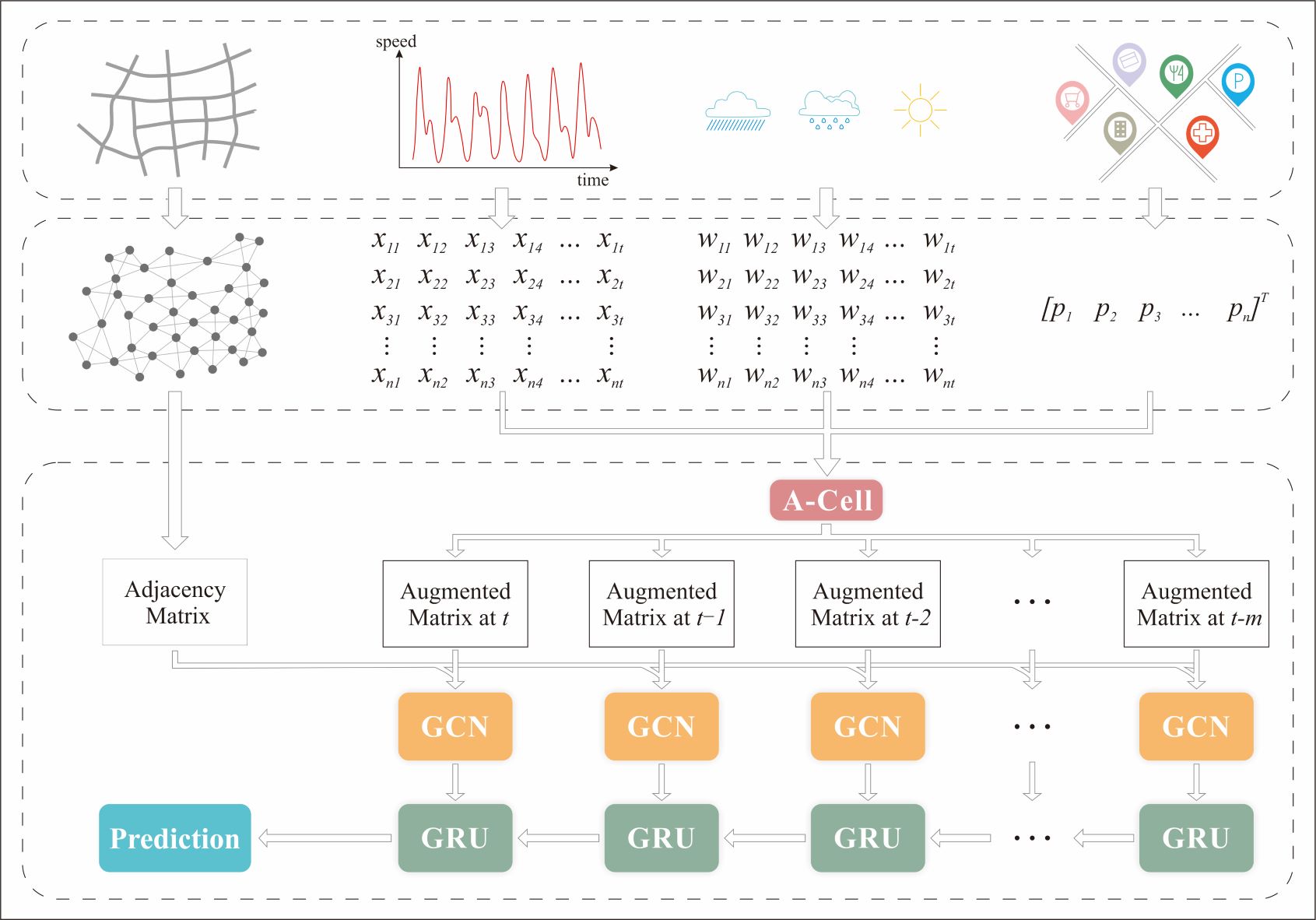}
     \caption{Framework.}
     \label{fig:6}
\end{figure}

By integrating the spatiotemporal graph convolution network and the attribute augmentation unit, we propose a traffic prediction model (AST-GCN) that can integrate the external influence information (weather condition and surrounding POIs in experiments) to facilitate traffic forecasting. The framework of our work is shown in Fig. \ref{fig:6}, which is mainly divided into four parts: data preprocessing, attribute augmentation, spatial-temporal dependency modeling and prediction.

\section{Experiments}
\subsection{Datasets}
The following datasets are used in the experiment:

(1) SZ\_taxi: this dataset contains Shenzhen taxi trajectory data collected from January 1 to January 31, 2015. A total of 156 major road sections in the Luohu District are selected, and their connectivity is modeled by a 156*156 adjacency matrix. The traffic speed time series of selected sections are calculated and form a feature matrix, where rows are indexed by road sections and columns are indexed by the timestamps.

(2) SZ\_POI: this dataset provides information about POIs surrounding selected road sections. The POI categories can be divided into 9 types: catering services, enterprises, shopping services, transportation facilities, education services, living services, medical services, accommodations, and others. After calculating the distribution of POIs on each road section, the type of POI with the largest proportion is used as the feature of the road section. Therefore, the obtained static attribute matrix is of size 156*1.

(3) SZ\_Weather: this auxiliary information contains the weather conditions about the study area recorded every 15 minutes in January 2015. The weather conditions are divided into five categories: sunny, cloudy, fog, light rain and heavy rain. With the information of time-varying weather conditions, we construct the dynamic attribute matrix with size 156*2,976.

\subsection{Evaluation metrics}
To evaluate the prediction performance of the proposed model, we use the following metrics to evaluate the prediction results.

(1) Root mean square error (RMSE)

 The smaller the RMSE value, the smaller the prediction error, and the better the performance of the model.

 \begin{equation}
  \begin{aligned}
    R M E S=\left[\frac{1}{n} \sum_{t=1}^{n}\left(y_{t}-\hat{y}_{t}\right)^{2}\right]^{\frac{1}{2}}
      \end{aligned}
  \label{equa:11}
 \end{equation}

(2) Mean absolute error (MAE)

MAE describes the average of the sum of the absolute difference between the predicted result and the ground truth. It is mainly used to evaluate the prediction error.

\begin{equation}
  \begin{aligned}
    M A E=\frac{\sum_{i=1}^{n}\left|y_{t}-\hat{y}_{t}\right|}{n}
      \end{aligned}
  \label{equa:12}
 \end{equation}

(3) Accuracy
\begin{equation}
  \begin{aligned}
    {Accuracy} =1-\frac{\|y-\hat{y}\|_{F}}{\|y\|_{F}}
        \end{aligned}
  \label{equa:13}
 \end{equation}

where $\|\cdot\|_{F}$ represents the Frobenius norm.

(4)($R^2$)

R-square is mainly used to measure the predictive ability of the model, and the larger the $R^2$, the better the prediction results.
\begin{equation}
  \begin{aligned}
    R^{2}=1-\frac{\sum_{t=1}\left(y_{t}-\hat{y}_{t}\right)^{2}}{\sum_{t=1}\left(y_{t}-\bar{y}_{t}\right)^{2}}
        \end{aligned}
  \label{equa:14}
 \end{equation}
(5) Explained variation (VAR)

VAR measures the proportion to which the proposed model accounts for the variation in real traffic states. It is mainly used to measure the predictive ability of the model.
\begin{equation}
  \begin{aligned}
    {Var}=1-\frac{{Var}(y-\hat{y})}{{Var}(y)}
  \end{aligned}
  \label{equa:15}
 \end{equation}

 \subsection{Parameter Settings}
The hyperparameters of the AST-GCN model mainly include the learning rate, training epoch, hidden units, batch size, and proportion of the training set. While the learning rate, batch size and proportion of the training set are manually set to 0.001, 64 and 0.8, the other parameters are searched through experiments.

\begin{figure}[ht]
	\centering
     \includegraphics[width=0.8\linewidth]{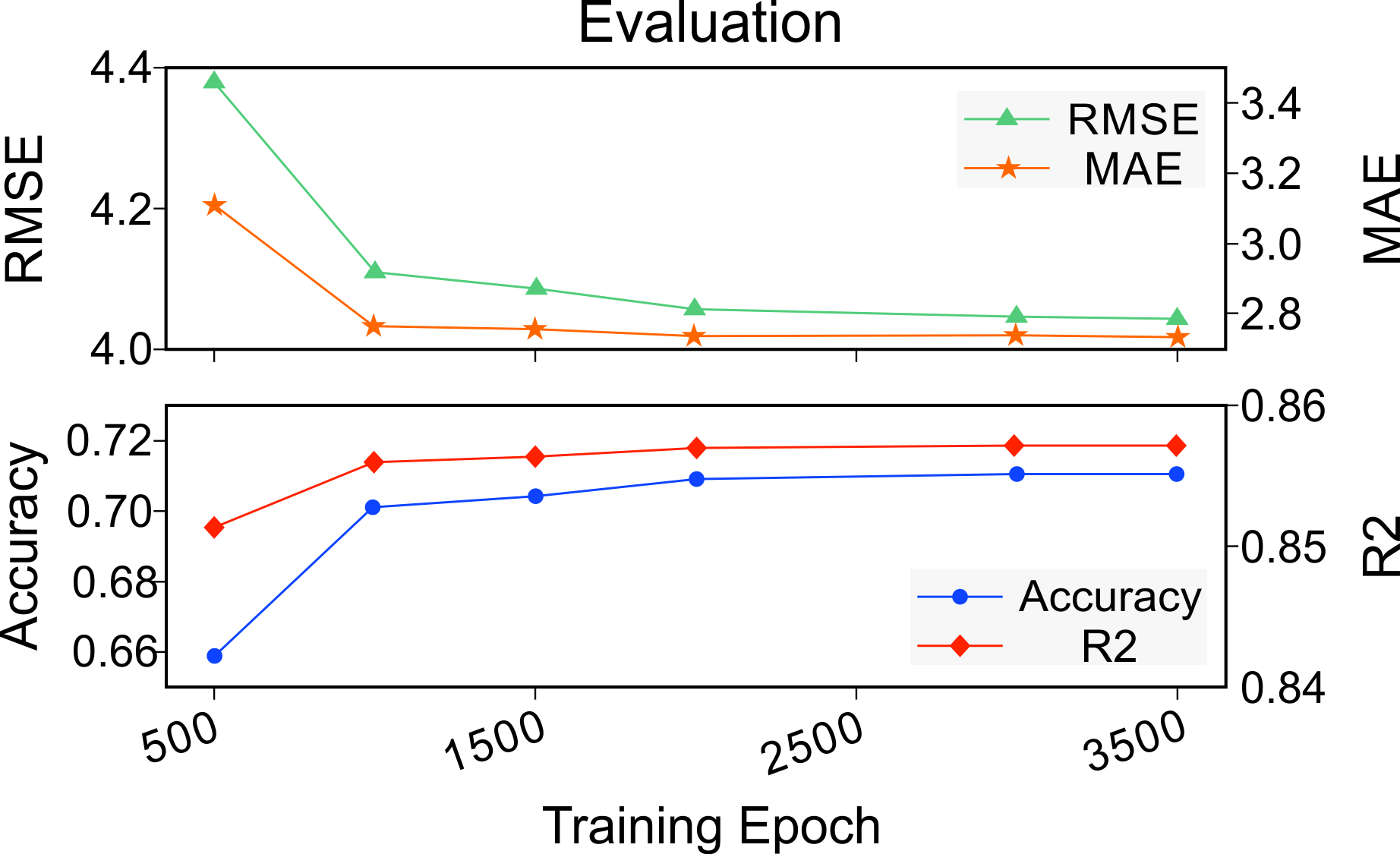}
     \caption{The influence of the selection of epochs on the prediction performance.}
     \label{fig:7}
\end{figure}

\begin{figure}[ht]
	\centering
     \includegraphics[width=0.8\linewidth]{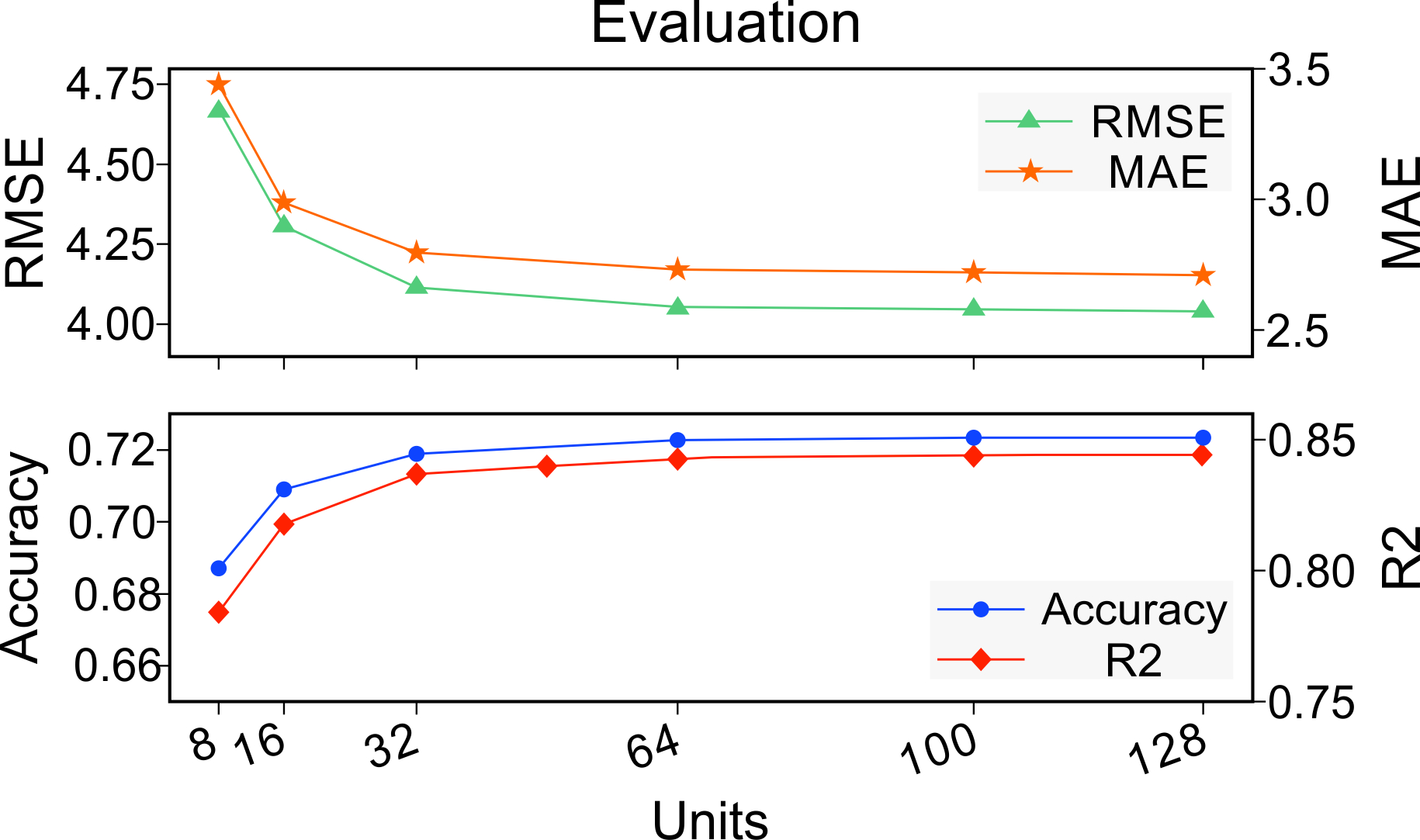}
     \caption{The influence of the selection of units on the prediction performance.}
     \label{fig:8}
\end{figure}

First, training epochs in set [500, 1,000, 1,500, 2,000, 3,000, 3,500] are tested to analyze the change in model performance. Fig. \ref{fig:7} shows the evaluation results under varying training epoch settings. As the value of the training epoch increases, the changes in the evaluation metrics become stable, and the turning point is 3,000. Then, when the training epoch is fixed to 3,000, the number of hidden units is selected from the candidate set [8, 16, 32, 64, 100, 128]. As shown in Fig. \ref{fig:8}, when the number of units reaches 100, the model becomes stable. Therefore, the training epoch is determined to be 3,000, while the number of hidden units is 100.


\subsection{Baselines}
We compare the proposed AST-GCN with the following baselines: (1) autoregressive integral moving average model (ARIMA)\cite{hamed1995short}, (2) support vector regression (SVR)\cite{su2007short}, (3) graph convolution model (GCN)\cite{kipf2016semi}, (4) gated recurrent unit model (GRU)\cite{chung2014empirical}, (5) spatiotemporal graph convolution model (TGCN)\cite{zhao2019t} and (6) diffusion convolutional recurrent neural network (DCRNN)\cite{li2017diffusion}. The hyperparameters of these baseline methods are the same as those in the original paper or released codes.

\subsection{Experimental Results}
Limited by the data sources, this paper simply uses the POI distributions and weather conditions to exemplify the importance of enhancing the model's ability to perceive external factors. The experiments are designed from five perspectives: the comparison of forecasting accuracy with baselines, the influence of introducing different kinds of external information, the influence of different predicting lengths, the interpretability of the proposed model, and the model’s robustness.

\subsubsection{Forecasting Comparison with Baselines}

To verify the effectiveness of the AST-GCN model in the traffic forecasting task, the prediction accuracy is compared with baselines, as shown in Table. \ref{tab:1}.

\begin{table*}[ht]
  \caption{Performance comparison of different models for traffic forecasting on the Shenzhen dataset.}
  \centering
  \begin{threeparttable}

  \begin{tabular}{cccccccc}
  \hline
  Evaluation metrics & SVR    & ARIMA  & GCN    & GRU    & DCRNN  & TGCN   & AST-GCN \\ \hline
  RMSE               & 7.2203 & 6.7708 & 5.6419 & 5.0649 & 4.5000 & 4.0696 & 4.0294  \\ 
  MAE                & 4.7762 & 4.6656 & 4.2265 & 2.5988 & 3.1700 & 2.7460 & 2.7035  \\
  Accuracy           & 0.7060 & 0.3852 & 0.6119 & 0.7243 & 0.2913 & 0.7165 & 0.7193  \\
  R2                 & 0.8367 & *      & 0.6678 & 0.8322 & 0.8391 & 0.8388 & 0.8512  \\
  var                & 0.8375 & 0.0111 & 0.6679 & 0.8322 & 0.8391 & 0.8388 & 0.8512 \\ \hline
  \end{tabular}
  \begin{tablenotes}
    \item * represents a negative value.
    \end{tablenotes}
\end{threeparttable}

  \label{tab:1}
\end{table*}

From the experimental results, it can be found that the prediction accuracy of the methods based on deep learning methods (AST-GCN, GCN, GRU, and TGCN) is higher than that of other methods. Compared with the SVR and ARIMA models, the RMSE of the attribute-aware AST-GCN model is reduced by approximately 44.19\% and 40.49\%, respectively. From the spatiotemporal perspective, compared with the GCN and GRU, which only focus on spatial or temporal relationships, the RMSE of AST-GCN, which considers both, is reduced by approximately 28.58\% and 20.44\%, respectively, and the overall evaluation given by other metrics is also significantly improved. From the perspective of attribute enhancement, the AST-GCN model, which considers external factors, outperforms spatiotemporal models such as TGCN and DCRNN, and the RMSE is reduced by approximately 0.98\% and 10.46\%, respectively. The comparison results verify the effectiveness of the proposed AST-GCN model.

\subsubsection{Ablation Experiments}

We conduct ablation experiments to prove that dynamic and static auxiliary information can play a role in the task of traffic forecasting. The experimental setting is divided into adding only static external information, adding only dynamic external information, adding dynamic and static external information, and not adding any external information. The results are shown in Table. \ref{tab:2}, and the fourth column is the result of adding dynamic weather conditions, the fifth column is the result of adding static POI distribution information, and the sixth column is the result of adding both.
\begin{table*}
  \caption{Ablation Experiments under different experimental settings.}

  \centering
  \begin{tabular}{cccccc} 
  \hline
  \multirow{2}{*}{Evaluation metrics} & \multirow{2}{*}{TGCN} & \multirow{2}{*}{DCRNN} & \multicolumn{3}{c}{AST-GCN}     \\ 
  \cline{4-6}
                                      &                       &                        & Weather & POI    & Weather+POI  \\ 
  \cline{1-3}\cline{4-6}
  RMSE                                & 4.0696                & 4.5000                 & 4.0378  & 4.0418 & 4.0294       \\
  MAE                                 & 2.7460                & 3.1700                 & 2.7059  & 2.6985 & 2.7035       \\
  Accuracy                            & 0.7165                & 0.2913                 & 0.7187  & 0.7180 & 0.7193       \\
  R2                                  & 0.8388                & 0.8391                 & 0.8506  & 0.8498 & 0.8512       \\
  Var                                 & 0.8388                & 0.8391                 & 0.8506  & 0.8498 & 0.8512       \\
  \hline
  \end{tabular}
  \label{tab:2}

  \end{table*}

From the perspective of simply adding dynamic external information, the RMSE of AST-GCN (weather) is reduced by approximately 0.78\% and 10.27\% compared with the TGCN and DCRNN models. From the perspective of simply adding static external information, the RMSE of AST-GCN (POI) is 0.68\% and 10.18\% lower than that of the TGCN and DCRNN models. After adding both dynamic and static external information, the RMSE of AST-GCN (weather+POI) is 0.98\% and 10.46\% lower than those of the TGCN and DCRNN models.

When considering the type of information to introduce, the model enhanced with dynamic external information is better than the model with static external information, indicating that the impact of weather conditions on traffic states is greater than that of the surrounding POIs. In addition, adding both dynamic and static external information, the model performs better than the model that adds a single type of information. The prediction error is lower by 0.21\% and 0.31\%, indicating the complementarity of dynamic and static factors. In general, enhancing the model with external information can facilitate the traffic forecasting task.

\subsubsection{The Forecasting Performance of AST-GCN for Different Predicting Horizons}

This experiment tests the performance of AST-GCN for different prediction horizons (15 min, 30 min, 45 min, and 60 min), and the performance comparison between the AST-GCN model and the baseline models under different prediction horizons is shown in Table. \ref{tab:3}. For the prediction horizon of 15 minutes, the RMSE of the AST-GCN model is approximately 0.98\% and 10.46\% lower than that of TGCN and DCRNN. For the prediction horizon of 30 minutes, the RMSE of the AST-GCN model's prediction error is approximately 0.59\% and 11.77\% lower than those of TGCN and DCRNN. For the prediction horizon of 45 min, the RMSE of the AST-GCN model is approximately 0.52\% and 11.26\% lower than that of TGCN and DCRNN, respectively. When the prediction horizon is set to 60 min, the RMSE of the AST-GCN model is approximately 0.64\% and 11.64\% lower than TGCN and DCRNN, respectively. It can be concluded that the AST-GCN model can maintain good performance for different horizons and the capability for long-term forecasting.
\begin{table}
  \caption{Performance comparison between the AST-GCN model and the baseline models for different prediction horizons.}
  \centering
  \begin{tabular}{cccccc} 
\cline{1-5}
  \multirow{2}{*}{Time}  & \multirow{2}{*}{Metric} & \multirow{2}{*}{TGCN} & \multirow{2}{*}{DCRNN} & \multirow{2}{*}{AST-GCN} &   \\
                         &                         &                       &                        &                          &   \\ 
 \cline{1-5}
  \multirow{5}{*}{15 min} & RMSE                    & 4.0696                & 4.5000                 & 4.0294                   &   \\
                         & MAE                     & 2.7460                & 3.1700                 & 2.7035                   &   \\
                         & Accuracy                & 0.7165                & 0.2913                 & 0.7193                   &   \\
                         & R2                      & 0.8388                & 0.8391                 & 0.8512                   &   \\
                         & Var                     & 0.8388                & 0.8391                 & 0.8512                   &   \\
  \multirow{5}{*}{30 min} & RMSE                    & 4.0770                & 4.5600                 & 4.0529                   &   \\
                         & MAE                     & 2.7470                & 3.2300                 & 2.7265                   &   \\
                         & Accuracy                & 0.7159                & 0.2970                 & 0.7176                   &   \\
                         & R2                      & 0.8377                & 0.8332                 & 0.8494                   &   \\
                         & Var                     & 0.8377                & 0.8360                 & 0.8495                   &   \\
  \multirow{5}{*}{45 min} & RMSE                    & 4.1035                & 4.6000                 & 4.0822                   &   \\
                         & MAE                     & 2.7788                & 3.2700                 & 2.7611                   &   \\
                         & Accuracy                & 0.7141                & 0.3021                 & 0.7156                   &   \\
                         & R2                      & 0.8357                & 0.8275                 & 0.8473                   &   \\
                         & Var                     & 0.8357                & 0.8314                 & 0.8474                   &   \\
  \multirow{5}{*}{60 min} & RMSE                    & 4.1266                & 4.6400                 & 4.1001                   &   \\
                         & MAE                     & 2.7911                & 3.3100                 & 2.7744                   &   \\
                         & Accuracy                & 0.7125                & 0.3069                 & 0.7143                   &   \\
                         & R2                      & 0.8339                & 0.8219                 & 0.8459                   &   \\
                         & Var                     & 0.8340                & 0.8267                 & 0.8460                   &   \\
  \cline{1-5}
  \end{tabular}
  \label{tab:3}
  \end{table}

\subsubsection{Perturbation Analysis}
\begin{figure}
  \centering
  \subfigure[Gaussian Perturbation]{
      \begin{minipage}{0.42\textwidth}
      \includegraphics[width=1\textwidth]{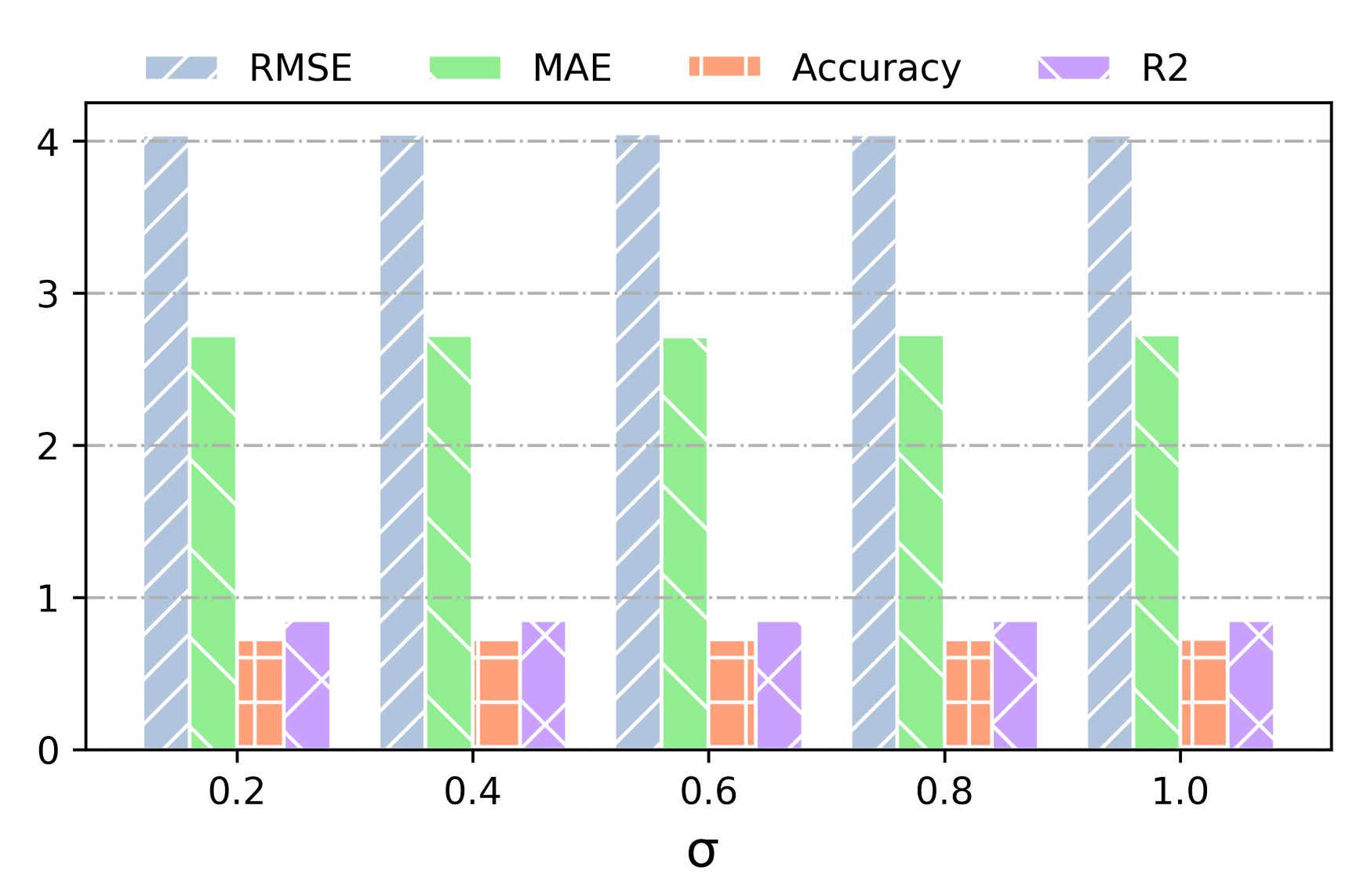}
      \end{minipage}
  }
  \subfigure[Poisson Perturbation]{
      \begin{minipage}{0.42\textwidth}
      \includegraphics[width=1\textwidth]{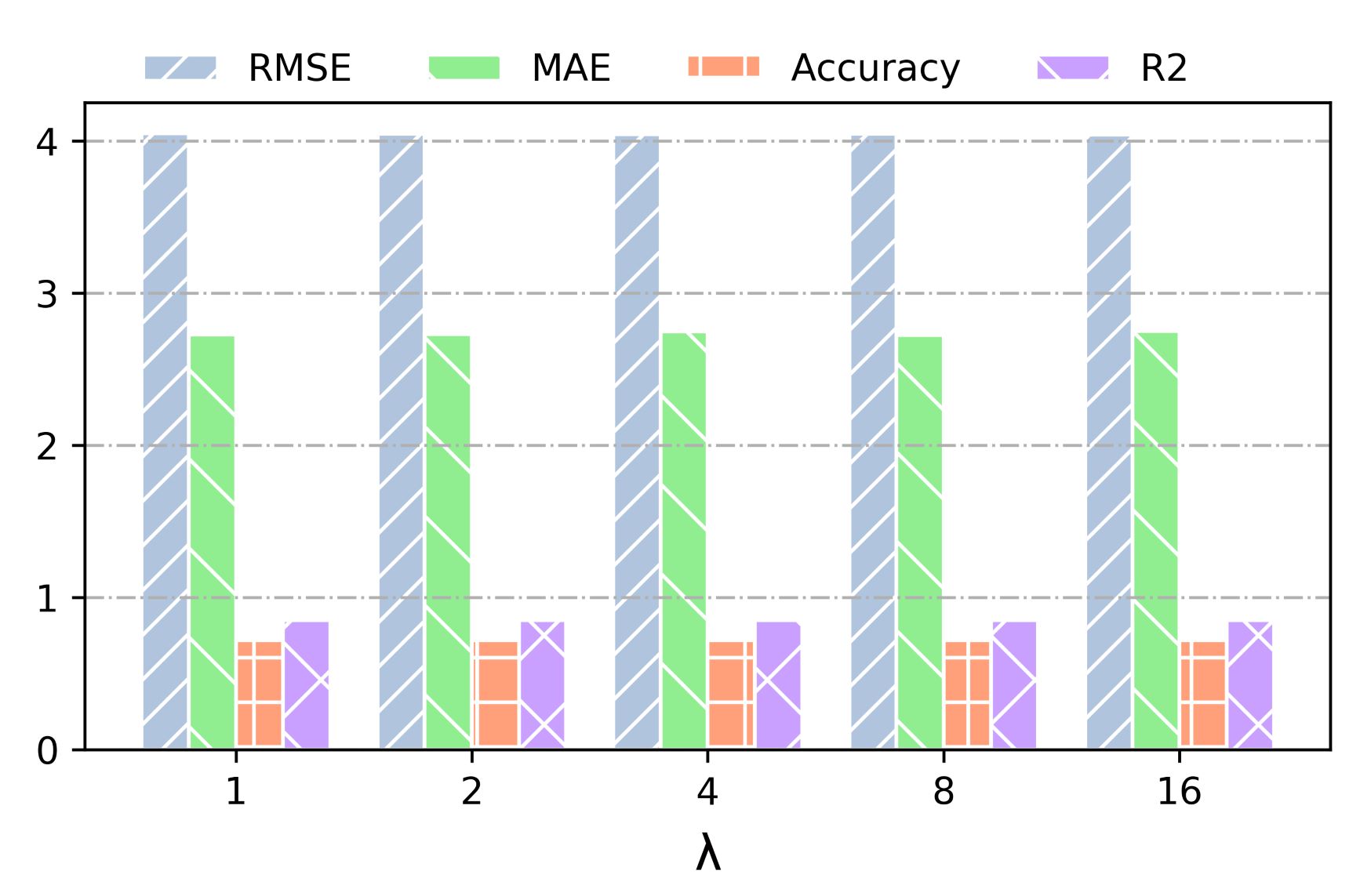}
      \end{minipage}
  }
  \caption{Perturbation Analysis.
  } \label{fig:19}
  \end{figure}

Gaussian noises and Poisson noises are added to the data to test the robustness of the model. The two types of noise obey Gaussian distribution $N \in\left(0, \sigma^{2}\right)(\sigma \in 0.2,0.4,0.6,0.8,1,2)$ and Poisson distribution $P(\lambda)(\lambda \in 1,2,4,8,16)$, respectively. The experimental results are shown in Fig. \ref{fig:19}, and the robustness of the AST-GCN model is verified by the fact that the changes in evaluation metrics across different noise settings are negligible.

\subsubsection{Interpretation of the AST-GCN}

To explain the predictive ability of the model more clearly, this experiment visually compares and analyze the true speed value of the test set and the prediction result of the attribute enhancement graph convolution model (AST-GCN) and explain the model from the following two perspectives:

(1) Long-term forecasting

\begin{figure}[ht]
	\centering
     \includegraphics[width=0.9\linewidth]{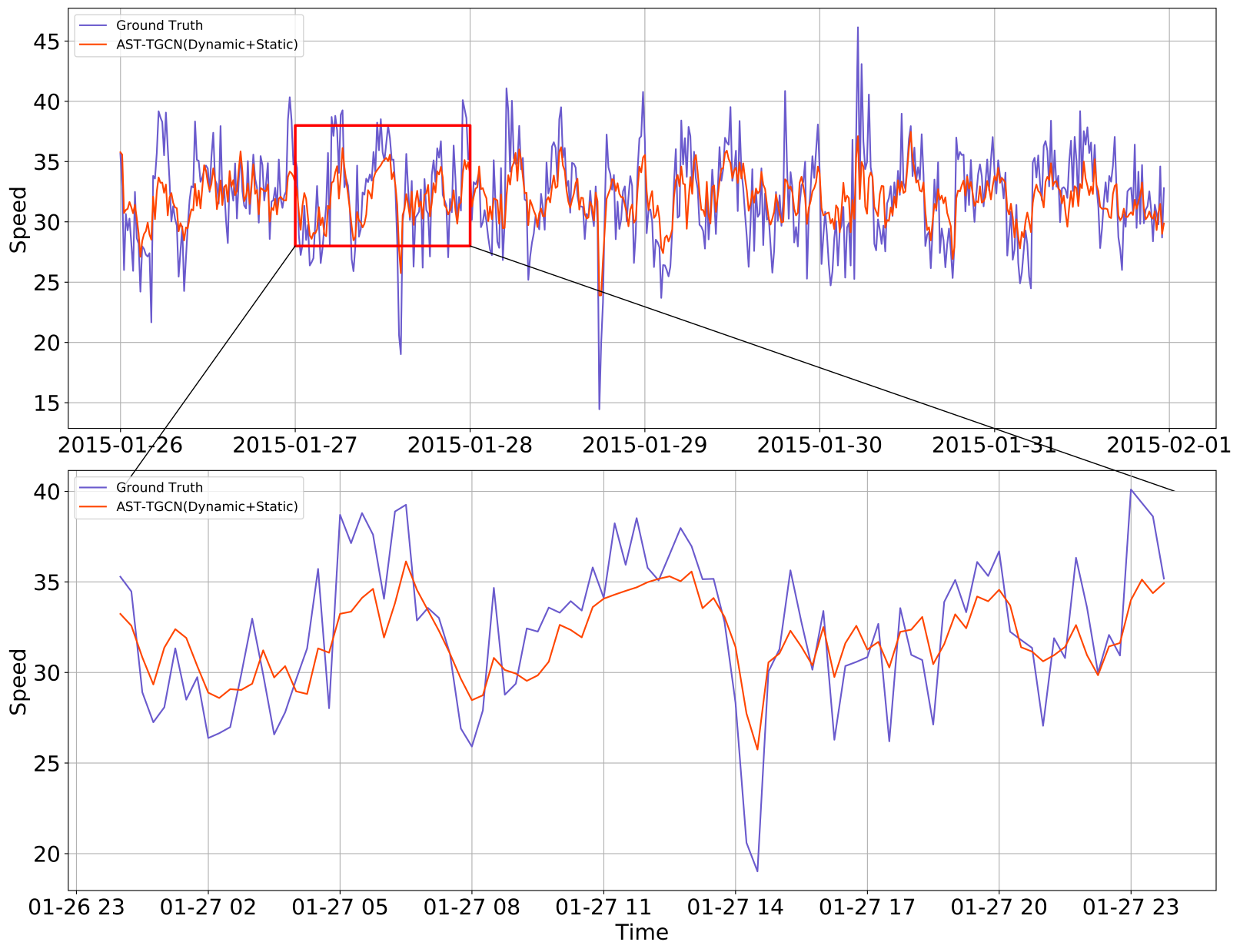}
     \caption{The visualization results for the 15 minute prediction horizen.}
     \label{fig:9}
\end{figure}

\begin{figure}[ht]
	\centering
     \includegraphics[width=0.9\linewidth]{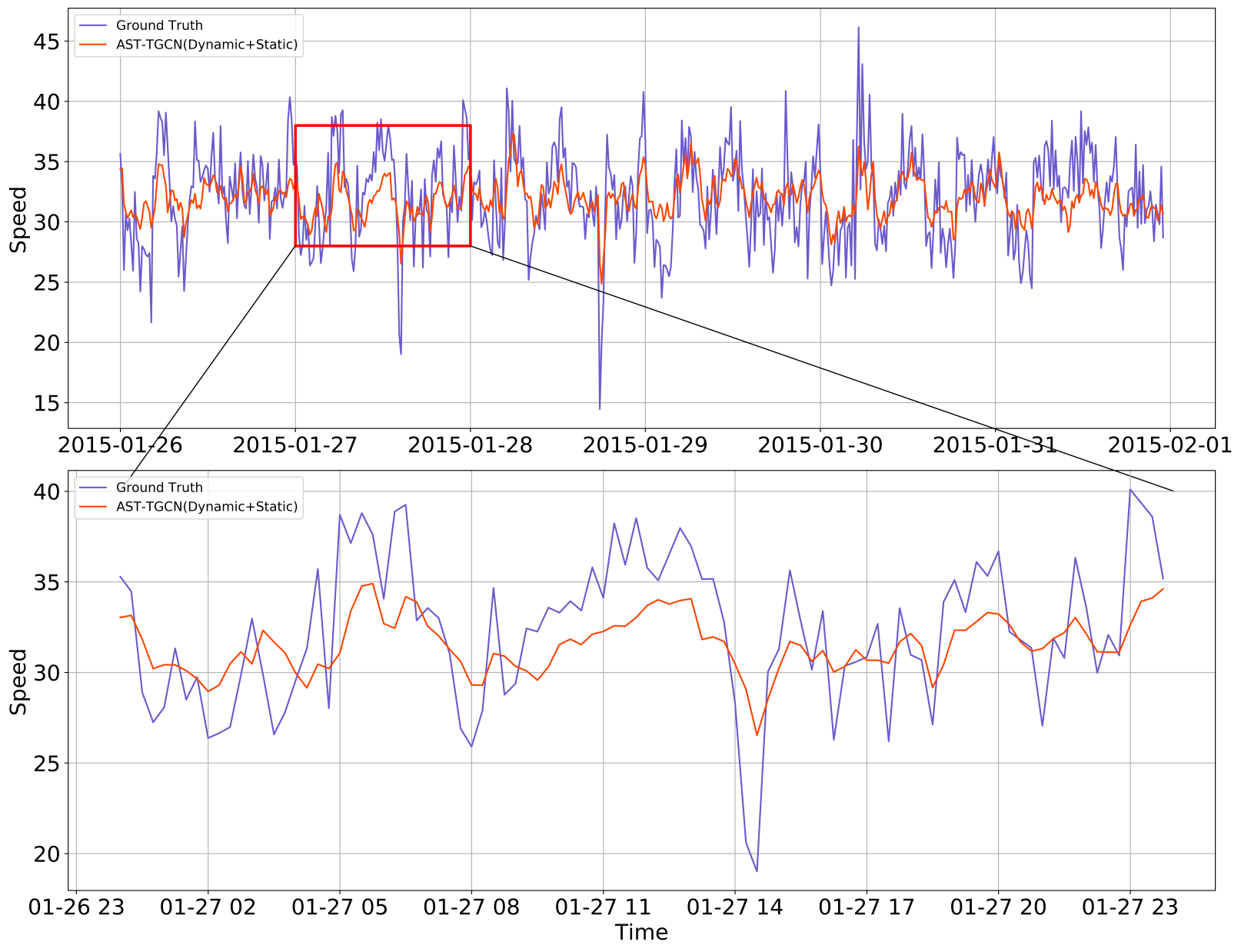}
     \caption{The visualization results for the 30 minute prediction horizon .}
     \label{fig:10}
\end{figure}

\begin{figure}[ht]
	\centering
     \includegraphics[width=0.9\linewidth]{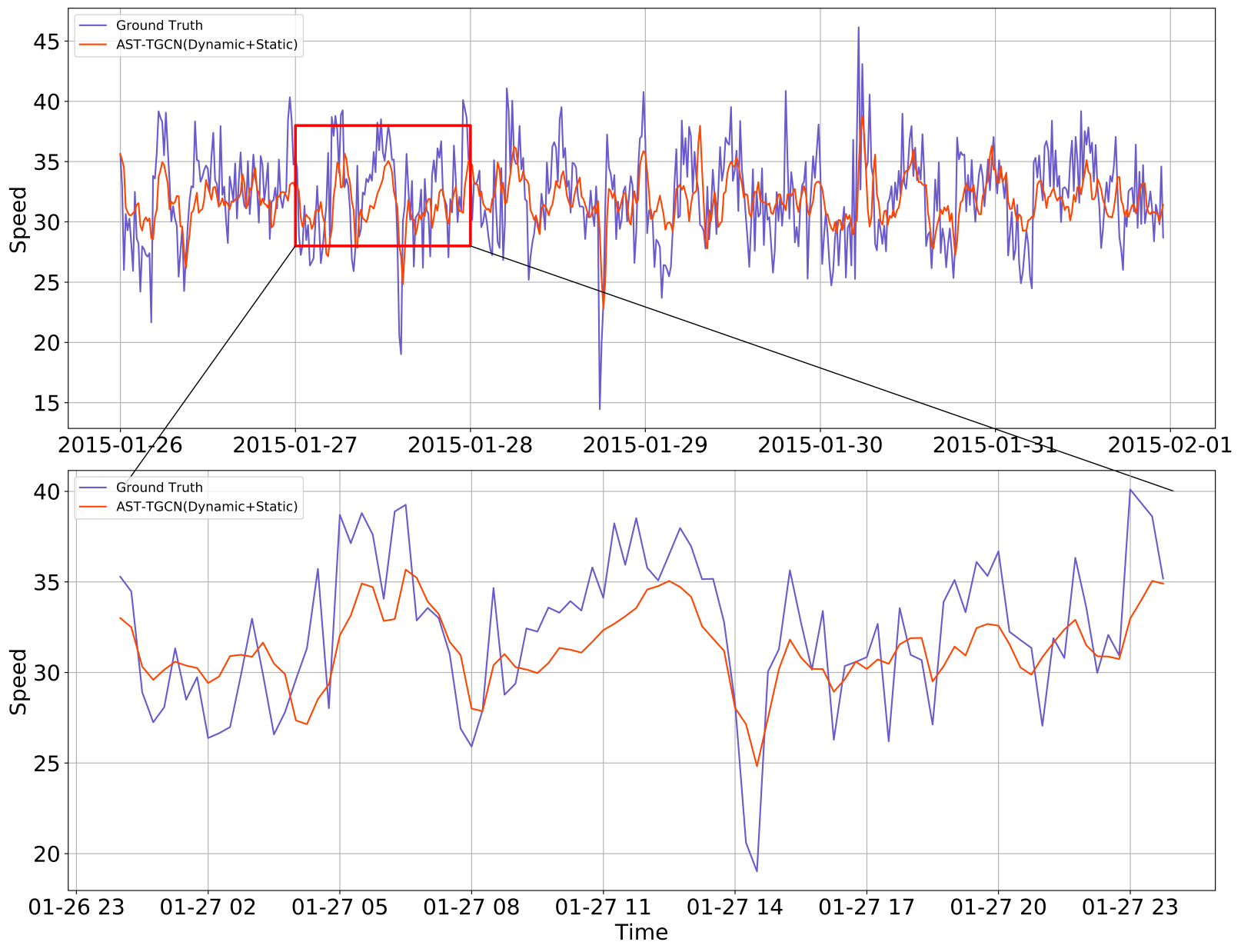}
     \caption{The visualization results for the 45 minute prediction horizon.}
     \label{fig:11}
\end{figure}

\begin{figure}[ht]
	\centering
     \includegraphics[width=0.9\linewidth]{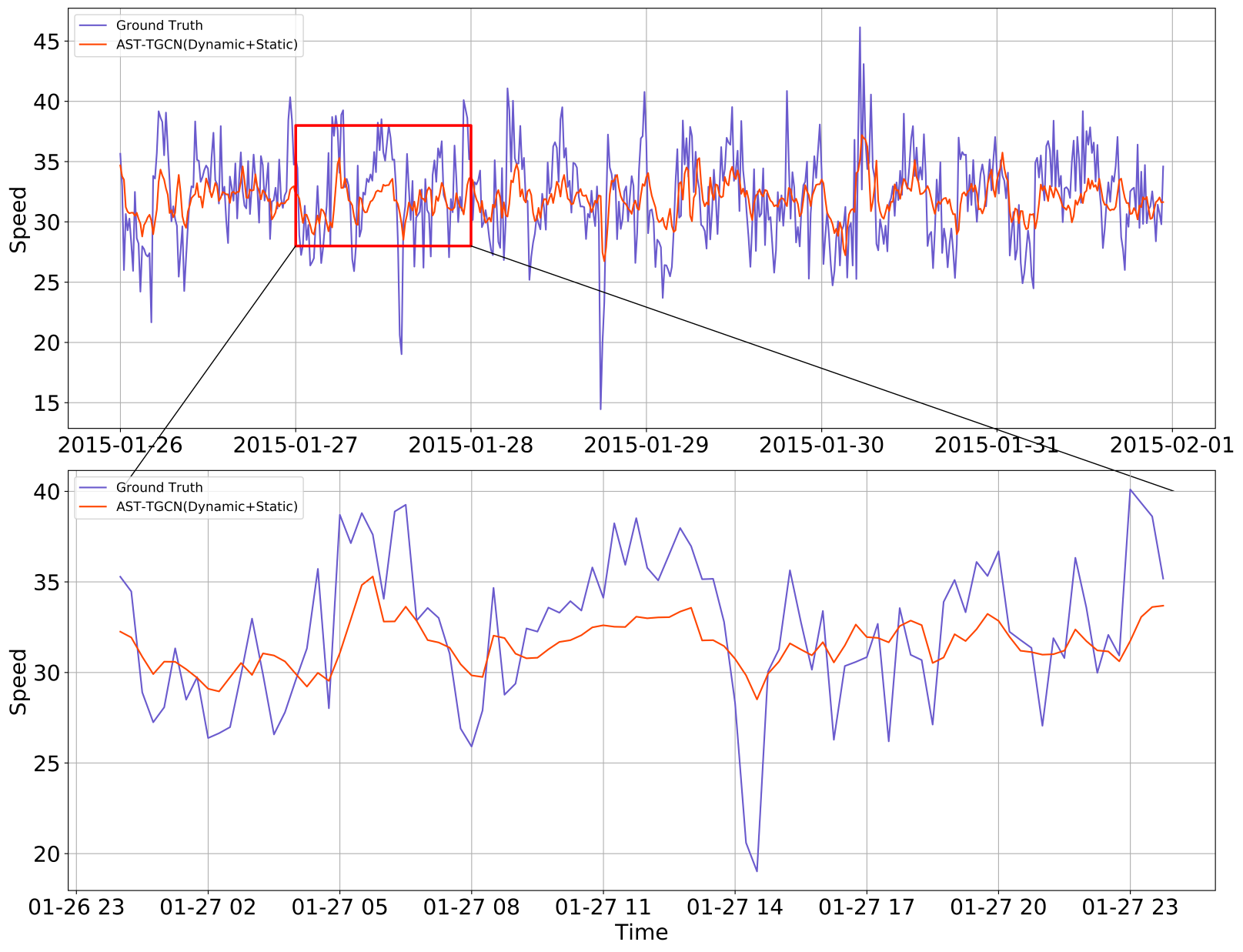}
     \caption{The visualization results for the 60 minute prediction horizon .}
     \label{fig:12}
\end{figure}

Based on the historical one-hour data, the results of predicting traffic speeds for the next 15 minutes, 30 minutes, 45 minutes, and 60 minutes are visualized in Fig. \ref{fig:9}-\ref{fig:12}.
The upper subfigure in each figure is the forecasting result from January 26, 2015, to January 31, 2015, and the lower image is the forecasting result for January 27, 2015. From the visualization results, we can conclude the following:
 \begin{itemize}
\item For different forecasting horizons, the model can well predict the traffic speed value. The prediction result of the model is similar to the changing trend of the real speed;
\item The performance of short-term forecasting is better than that of long-term forecasting. Comparing the prediction results for 15 min and 60 min, it can be found that the curve of short-term prediction is closer to the real curve, which indicates that the model can better capture the short-range dependencies while losing some information in long-term prediction;
\item For the capturing of the turning points of the speed changing trend, the AST-GCN model has considerable deviations at high and low peaks. The reason may be that the sudden change in traffic states is affected not only by the weather and POI factors used in this study but also by a combination of other factors.
\end{itemize}


(2) Attribute importance

\begin{figure}[ht]
  \centering
     \includegraphics[width=0.9\linewidth]{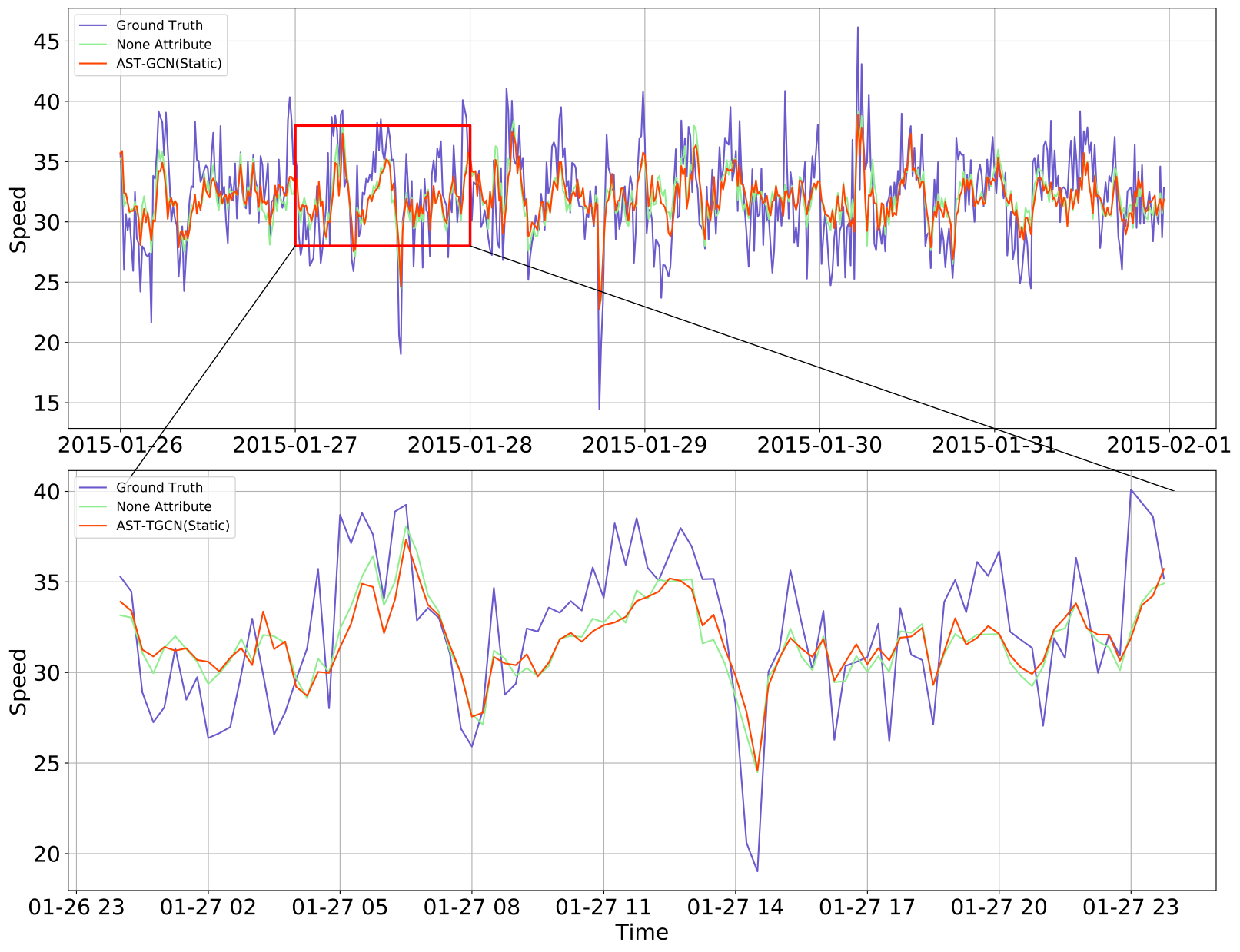}
     \caption{Comparison between forecasting enhanced by static POI information and forecasting without external information.}
     \label{fig:13}
\end{figure}

\begin{figure}[ht]
  \centering
     \includegraphics[width=0.9\linewidth]{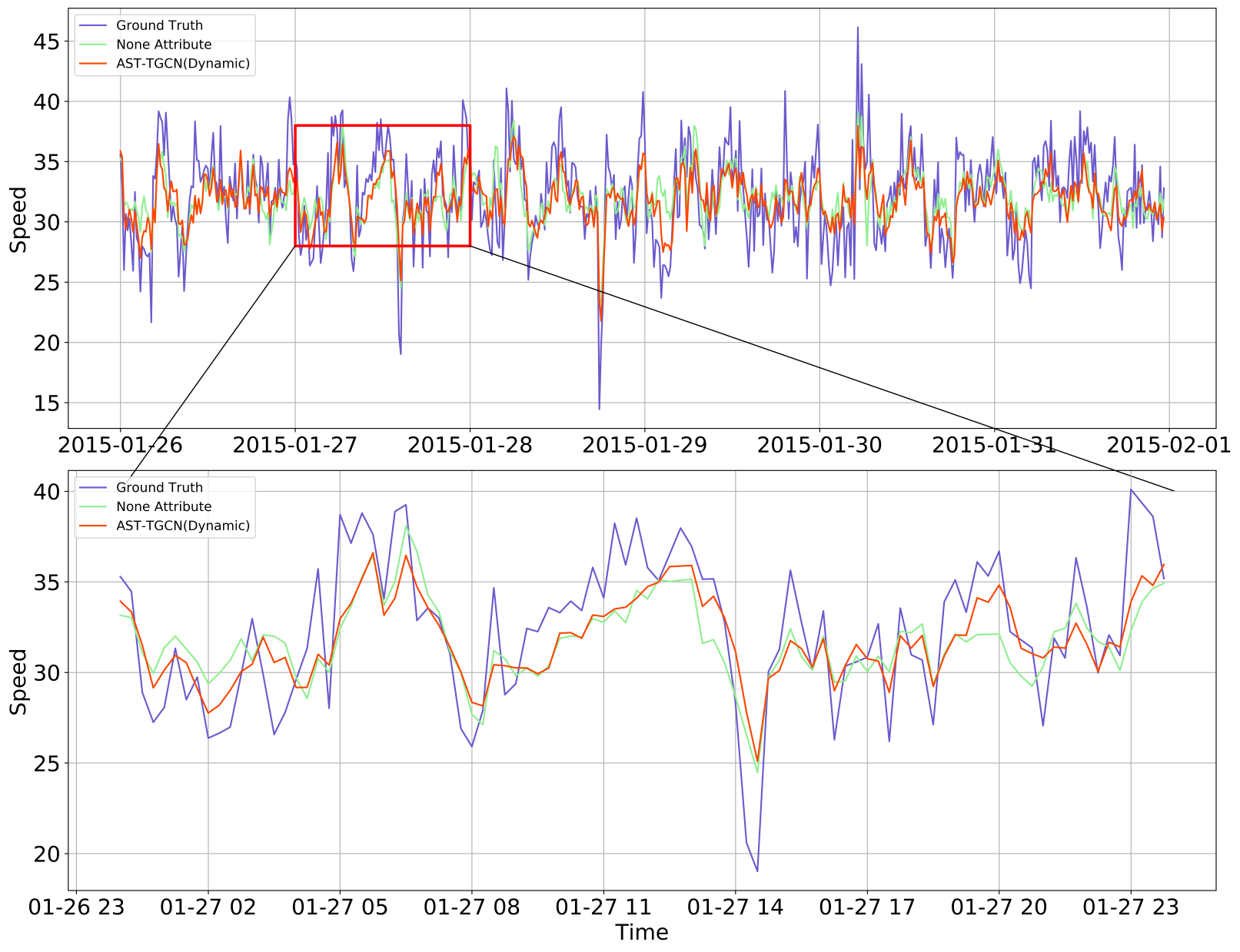}
     \caption{Comparison between forecasting enhanced by dynamic weather condition information and forecasting without external information.}
     \label{fig:14}
\end{figure}

\begin{figure}[ht]
  \centering
     \includegraphics[width=0.9\linewidth]{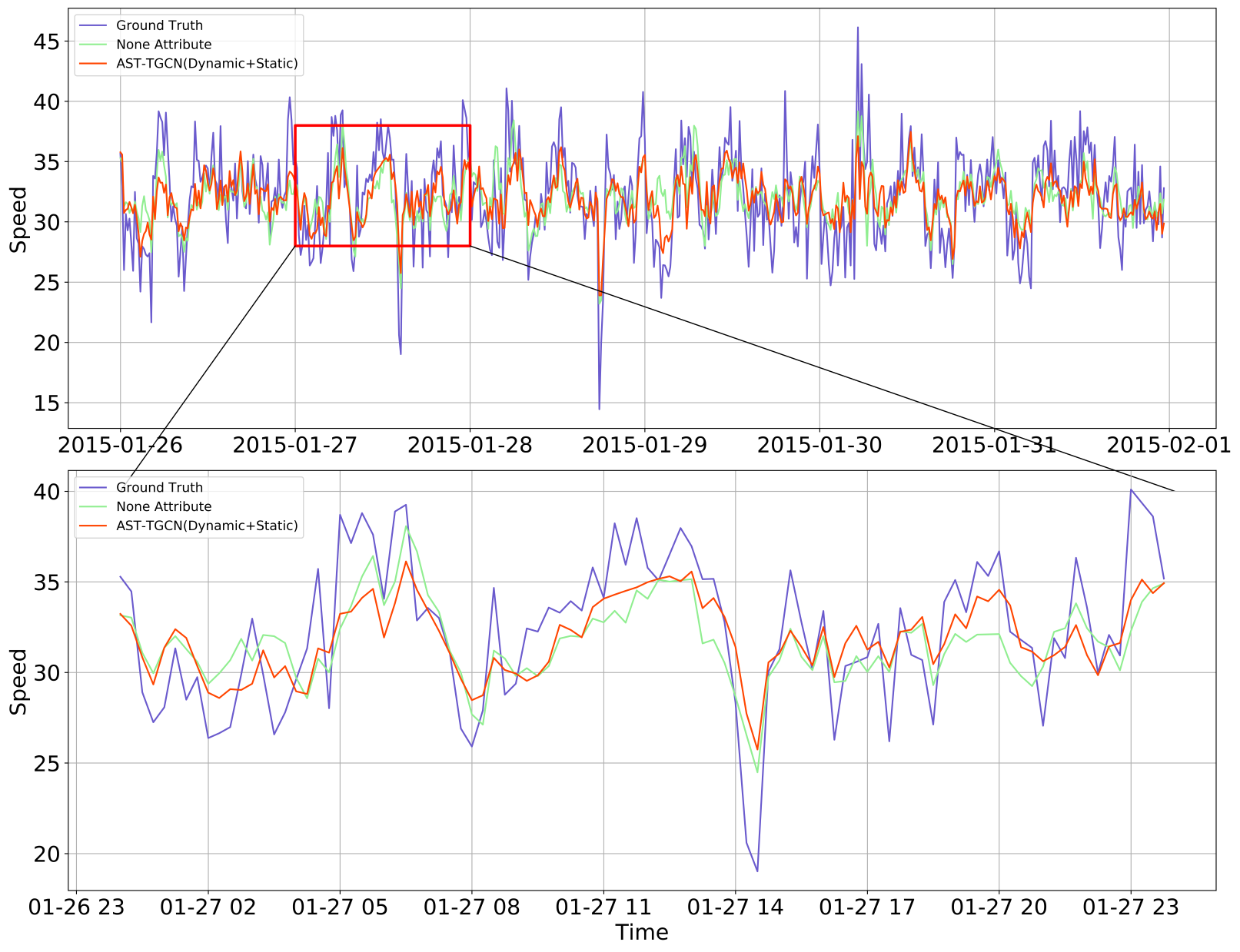}
     \caption{Comparison between forecasting enhanced by the combination of static and dynamic information and forecasting without external information.}
     \label{fig:15}
\end{figure}

\begin{figure}[ht]
  \centering
     \includegraphics[width=0.9\linewidth]{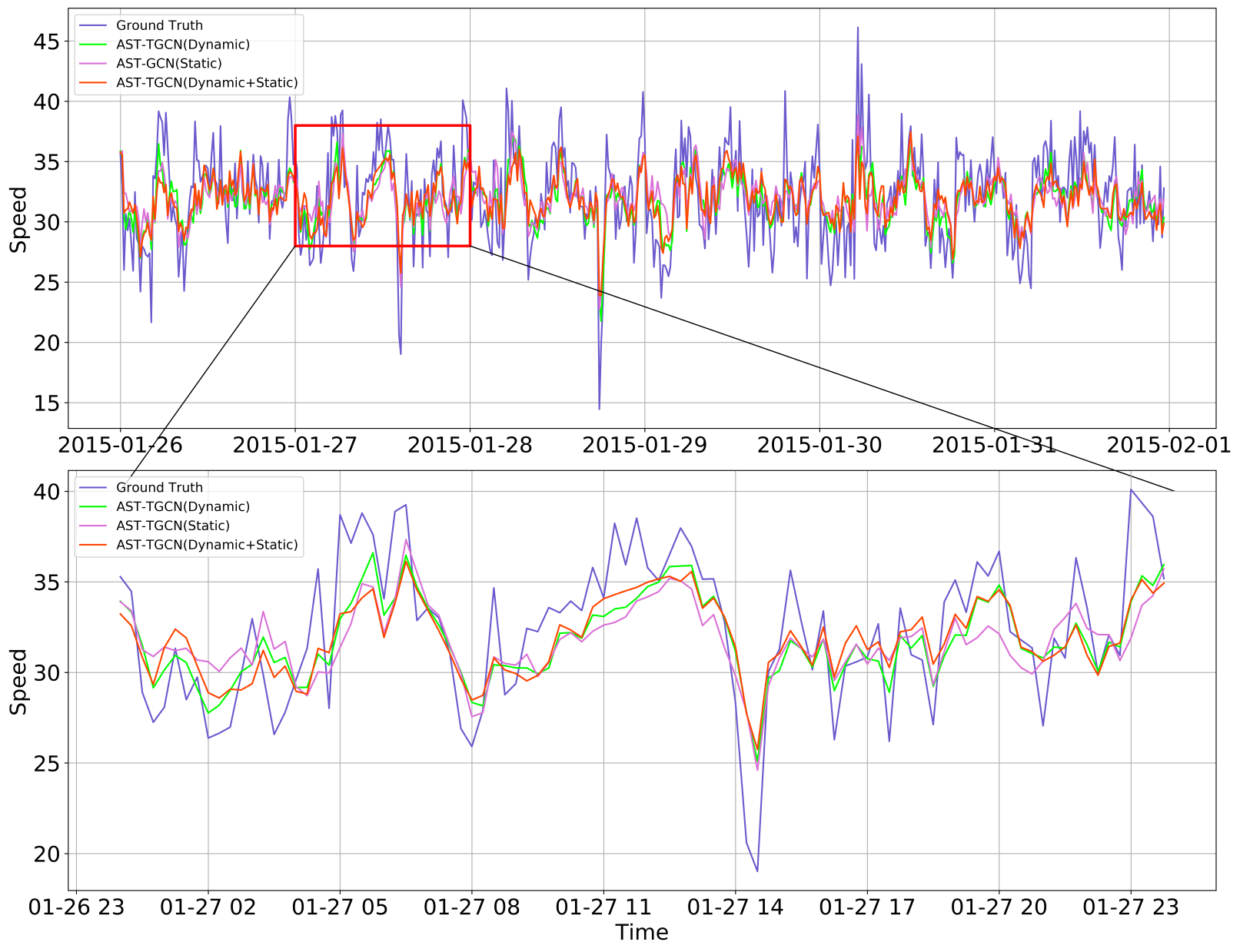}
     \caption{Comparison among forecasting enhanced by different external information.}
     \label{fig:16}
\end{figure}

To further study the validity of static and dynamic external information, we visualize the results of the ablation experiments. Fig. \ref{fig:13}-\ref{fig:15} show the visual comparison of forecasting enhanced by static POI information, dynamic weather condition information, and their combination with forecasting without external information. 
shows the comparison between the forecasting results enhanced by different attributes.
 \begin{itemize}
\item External information improves the model's perception of peaks and turning points. Fig. \ref{fig:13}-\ref{fig:15} show that the prediction results of the model with external information are closer to the ground truth than the prediction result without attribute auxiliary information at turning points and peaks.
\item From the visualization results in Fig. \ref{fig:16}, it can be found that the deviation between the predicted results of AST-GCN (dynamic+static attribute) and the real speed value is smaller than that of AST-GCN (static attribute) and AST-GCN (dynamic attribute), which indicates that the diversity of external information can better facilitate forecasting.
\end{itemize}


\section{Conclusion}
This paper addresses the problem that the traditional urban traffic forecasting models cannot comprehensively consider the external factors that affect the traffic states and proposes an attribute-enhanced spatiotemporal graph convolution model AST-GCN. The model can not only integrate static external information but also dynamic external data. Through comparison with the baseline methods, the prediction results verify the importance of considering external information in traffic forecasting tasks. In addition, this paper uses a perturbation analysis to test the robustness of the model.


%

\appendices


\ifCLASSOPTIONcompsoc
  \section*{Acknowledgments}
\else
  \section*{Acknowledgment}
\fi

This work was supported by the National Science Foundation of China [grant numbers ].

\ifCLASSOPTIONcaptionsoff
  \newpage
\fi



\bibliographystyle{IEEEtran}
\bibliography{AST-GCN}
\end{document}